\documentclass[10pt,a4paper]{article}
\usepackage[utf8]{inputenc}
\usepackage[margin=1.5in]{geometry}
\usepackage{graphicx}
\usepackage{authblk}
\usepackage{hyperref}
\hypersetup{
    colorlinks=true,
    linkcolor=blue,
    filecolor=magenta,
    urlcolor=cyan,
}
\usepackage{cite}
\usepackage{amsmath}
\usepackage{amsfonts}
\usepackage{booktabs}

\title{\textbf{Interactive Medical-SAM2 GUI: A Napari-based semi-automatic annotation tool for medical images}}

\author[1]{Woojae Hong}
\author[2]{Jong Ha Hwang}
\author[1]{Jiyong Chung}
\author[1]{Joongyeon Choi}
\author[1]{Hyunggun Kim\thanks{Corresponding authors: \texttt{hkim.bme@skku.edu}, \texttt{kimyh96@snu.ac.kr}}}
\author[2]{Yong Hwy Kim\protect\footnotemark[1]}

\affil[1]{Department of Biomechatronic Engineering, Sungkyunkwan University, Suwon, Gyeonggi, Republic of Korea}
\affil[2]{Department of Neurosurgery, Seoul National University Hospital, Seoul National University College of Medicine, Seoul, Republic of Korea}

\begin{document}
\date{} 
\maketitle

\begin{abstract}
    Interactive Medical-SAM2 GUI is an open-source desktop application for semi-automatic annotation of 2D and 3D medical images. 
    Built on the Napari multi-dimensional viewer, box/point prompting is integrated with SAM2-style propagation by treating a 3D volume as a slice sequence, enabling mask propagation from sparse prompts using Medical-SAM2 on top of SAM2. 
    Voxel-level annotation remains essential for developing and validating medical imaging algorithms, yet manual labeling is slow and expensive for 3D scans, and existing integrations frequently emphasize per-slice interaction without providing a unified, cohort-oriented workflow for navigation, propagation, interactive correction, and quantitative export in a single local pipeline.
    To address this practical limitation, a local-first Napari workflow is provided for efficient 3D annotation across multiple studies using standard DICOM series and/or NIfTI volumes.
    Users can annotate cases sequentially under a single root folder with explicit proceed/skip actions, initialize objects via box-first prompting (including first/last-slice initialization for single-object propagation), refine predictions with point prompts, and finalize labels through prompt-first correction prior to saving.
    During export, per-object volumetry and 3D volume rendering are supported, and image geometry is preserved via SimpleITK.
    The GUI is implemented in Python using Napari and PyTorch, with optional N4 bias-field correction, and is intended exclusively for research annotation workflows.
    The code is released on the project page: \url{https://github.com/SKKU-IBE/Medical-SAM2GUI/}.
\end{abstract}

\begin{figure}[h]
    \centering
    \includegraphics[width=1.0\textwidth]{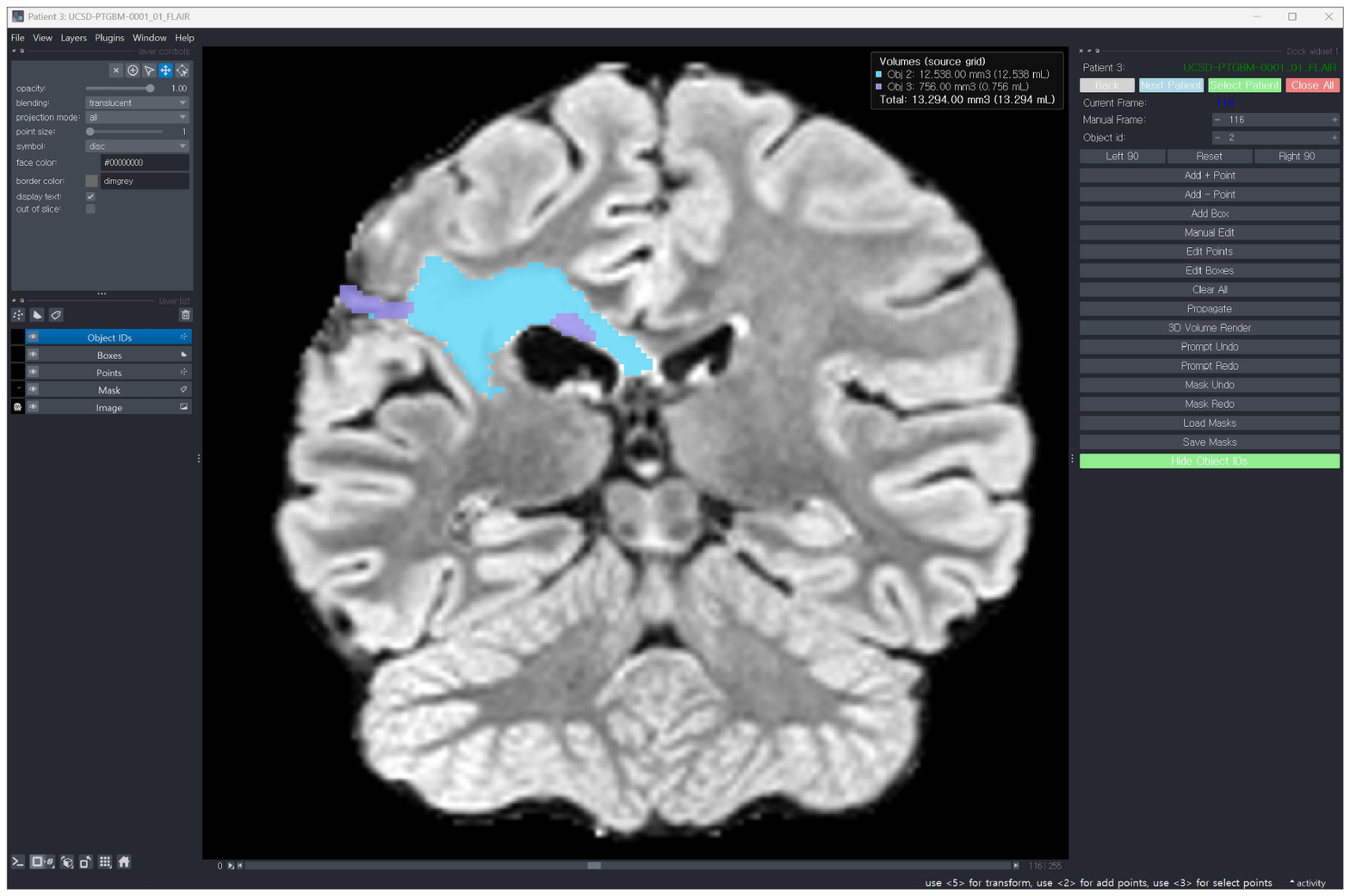}
    \caption{Primary view of the Interactive Medical-SAM2 GUI in Napari displaying an image, prompt layers, and mask overlays.}
    \label{fig:main_view}
\end{figure}

\section{Summary}
Interactive Medical-SAM2 GUI is an open-source desktop application for semi-automatic annotation of 2D and 3D medical images (Figure 1). 
Built on the Napari multi-dimensional viewer \cite{napari2025}, it integrates box/point prompting with SAM2-style propagation (treating a 3D scan as a “video” of slices) using Medical-SAM2 \cite{zhu2024medical} on top of SAM2 \cite{ravi2025sam2}. 
The tool is designed for clinician-friendly workflows: users can place DICOM series and/or NIfTI volumes under a single root folder (Figure 2a) and annotate cases sequentially, choosing to proceed or skip each case without repeatedly browsing individual patient files (Figure 2b). 
During saving, the tool reports per-object volumetry and provides 3D volume rendering to support rapid inspection and quantitative tracking (e.g., tumor burden) (Figure 2c).

Interactive Medical-SAM2 GUI is an open-source desktop application for semi-automated annotation of 3D medical image volumes (Figure 1). Built on the Napari multi-dimensional viewer \cite{napari2025}, it integrates box/point prompting with SAM2-style propagation (treating a 3D scan as a “video” of slices) using Medical-SAM2 \cite{zhu2024medical} on top of SAM2 \cite{ravi2025sam2}. The tool is designed for clinician-friendly workflows. Users can place DICOM series and/or NIfTI volumes under a single root folder (Figure 2a) and annotate cases sequentially, choosing to proceed or skip each case without repeatedly browsing individual patient files (Figure 2b). Existing multi-label masks can be reloaded for continuous annotation, and manual corrections are synchronized to the original image grid prior to exporting. While editing and saving, the tool provides color-matched per-object volumetry and optional 3D volume rendering to support rapid inspection and quantitative tracking such as tumor burden (Figure 2c).

\begin{figure}[h]
    \centering
    \includegraphics[width=1.0\textwidth]{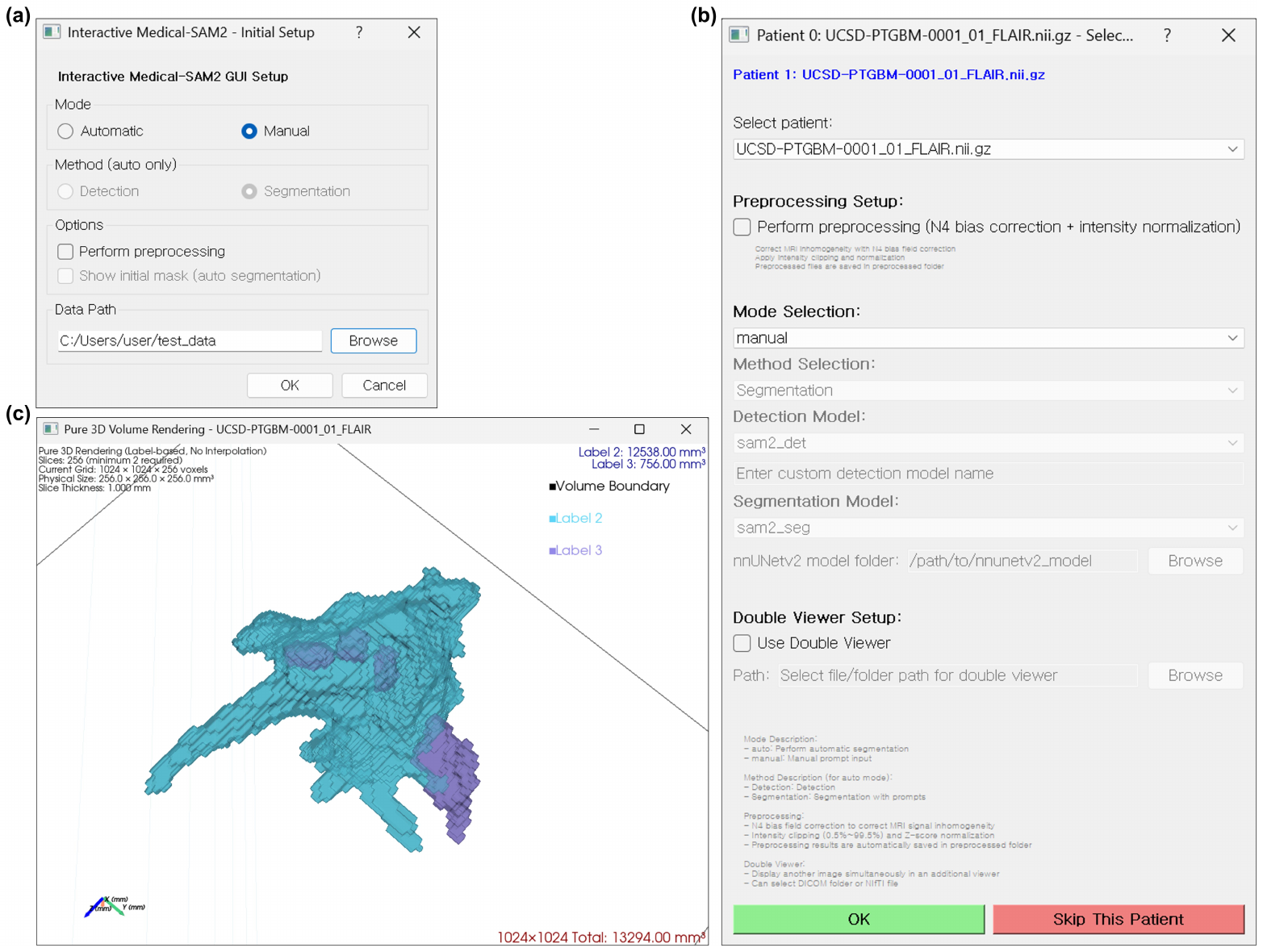}
    \caption{(a) Root-folder selection for DICOM and/or NIfTI data; (b) patient-by-patient navigation using proceed or skip; (c) 3D volume rendering that reports per-object volumetry computed from the saved masks.}
    \label{fig:workflow}
\end{figure}

\section{Statement of Need}
Voxel-level annotation is essential for developing and validating medical imaging algorithms, yet manual labeling is slow and expensive, especially for 3D scans containing hundreds of slices. Expert-friendly platforms such as ITK-SNAP \cite{yushkevich2006itksnap}, 3D Slicer \cite{fedorov2012slicer}, and MITK \cite{wolf2005mitk} provide robust visualization and classical semi-automatic segmentation tools. However, it still requires substantial manual work and careful data handling to produce consistent 3D labels at cohort scale.

AI-assisted labeling frameworks have improved annotation efficiency by combining model inference and active learning strategies. MONAI Label supports both local (3D Slicer) and web frontends and provides a comprehensive framework for deploying AI-driven annotation applications \cite{diazpinto2024monailabel}. While web-based labeling can be attractive for accessibility, clinical deployment is often constrained by institutional data governance and privacy requirements unless de-identification and secure hosting are rigorously validated, motivating local-first workflows for routine annotation. DeepEdit and similar interactive refinement approaches learn from simulated user edits to reduce the number of manual interactions necessary for generating accurate 3D segmentations \cite{diazpinto2022deepedit}.

Promptable foundation models have recently lowered the barrier to interactive segmentation. Segment Anything Model (SAM) \cite{kirillov2023sam} and its medical adaptations such as MedSAM \cite{ma2024medsam} have motivated integrations into standard annotation platforms, including 3D Slicer extensions (e.g., MedSAMSlicer \cite{medsamslicer2024}) and Napari plugins (e.g., napari-sam \cite{naparisam2023}). Medical-SAM2 extends SAM2's memory-based video segmentation approach to volumetric medical imaging by treating 3D volumes as slice sequences, thereby enabling segmentation propagation from sparse annotations across multiple slices \cite{zhu2024medical, ravi2025sam2}. However, existing integrations predominantly emphasize per-slice interaction and lack a unified, cohort-oriented workflow that seamlessly integrates navigation, propagation, interactive correction, and quantitative export within a single local pipeline.

Interactive Medical-SAM2 GUI addresses this practical limitation by integrating Medical-SAM2 propagation into a local-first Napari workflow designed for efficient 3D annotation across multiple patient studies using only standard DICOM or NIfTI inputs.

\section{State of the Field and Differentiation}

\subsection{General medical imaging workbenches}
3D Slicer and MITK offer extensive ecosystems of modules for segmentation, registration, and visualization \cite{fedorov2012slicer, wolf2005mitk}, while ITK-SNAP remains widely used for interactive 3D segmentation using user-guided active contour methods \cite{yushkevich2006itksnap}. Although these platforms are robust, repetitive annotation tasks may require additional tooling to ensure standardized navigation, prompt-based propagation, and consistent quantitative data export across large datasets.

\subsection{Interactive ML labeling tools and general annotators}
The interactive learning and segmentation toolkit (ilastik) provides interactive machine-learning workflows for segmentation, classification, and tracking that adapt to a task using sparse user annotations and supports data processing up to 5D \cite{berg2019ilastik}. In digital pathology, QuPath supports efficient annotation and scripting capabilities for large whole-slide images \cite{bankhead2017qupath}. While generic data-labeling platforms (e.g., CVAT \cite{cvat} and Label Studio \cite{labelstudio}) provide flexible web-based segmentation interfaces, these platforms typically require additional engineering modifications to handle medical imaging standards (DICOM/NIfTI), preserve geometric integrity, and support radiology-style workflows.

\subsection{Promptable foundation-model integrations}
Community integrations such as MedSAMSlicer [@medsamslicer2024] and napari-sam [@naparisam2023] have demonstrated strong demand for prompt-based labeling within established medical image viewers. Interactive Medical-SAM2 GUI adopts an alternative strategy to provide a single, clinician-oriented workflow: \textbf{navigation → prompting/propagation → final correction → quantitative export.}

\begin{enumerate}
    \item \textbf{Cohort navigation:} Users provide a single root path containing patient studies and annotate cases sequentially through explicit actions to proceed or skip, thereby minimizing manual file handling during routine labeling process. Generated masks and NIfTI label maps are excluded from the patient queue.
    \item \textbf{Box-first prompting and propagation:} Box prompts are the primary interaction for object initialization. For single-object annotation, users place box prompts on the first and last slices containing the target object, after which propagation generates masks for intermediate slices using Medical-SAM2.
    \item \textbf{Multi-object support with explicit control:} Multiple objects can be annotated within the same volume. In multi-object scenarios, prompts can be provided on relevant slices for each object to maintain user control in complex cases.
    \item \textbf{Point prompts for refinement:} Point prompts can be incorporated to refine slice-level predictions. In the current workflow, a bounding box prompt initially defines the object on a given slice, while point prompts provide additional guidance for small additions or corrections.
    \item \textbf{Prompt-first correction and resumable annotation:} Users typically obtain optimal segmentation through prompts and propagation, followed by final manual correction to “lock in” the label prior to saving. This workflow aligns with a propagation framework that operates primarily through prompt-based interactions, ensuring consistency and reproducibility. Previously saved or externally produced label maps can be loaded, aligned to the source geometry, and edited without discarding unaffected slices or objects.
    \item \textbf{Quantitative export and visualization:} Upon mask generation, the tool computes per-object volumetric measurements suitable for longitudinal tumor volume monitoring and provides 3D volume rendering capabilities for visual inspection of reconstructed anatomical structures. Saved masks preserve image geometry via SimpleITK \cite{lowekamp2013simpleitk}.
\end{enumerate}

\section{Software design}
The GUI is implemented in Python using Napari for multi-dimensional visualization [@napari2025] and PyTorch for model execution \cite{paszke2019pytorch}. Medical-SAM2 \cite{zhu2024medical} provides SAM2-style memory-based propagation across slice sequences \cite{ravi2025sam2}. Image I/O, geometry preservation (spacing, origin, and direction), and mask saving are handled primarily using SimpleITK \cite{lowekamp2013simpleitk}. For DICOM series with malformed or zero-valued slice-spacing metadata, pydicom \cite{mason2024pydicom} provides fallback recovery of slice order and spacing.

The editable Napari mask is maintained as display representation, whereas the canonical multi-label mask remains on the original image grid. This separation allows interactive editing and display rotation without modifying the source arrays or saved image geometry. Existing NIfTI, NRRD, and MetaImage label maps can be reloaded for continued annotation, and geometry mismatches are resampled using nearest-neighbor interpolation following user confirmation if needed. Volumetric measurements and geometry-preserving mask exports are derived from the same source-grid representation, thereby preventing discrepancies between the displayed annotation, reported volumes, and saved labels. This design supports resumable annotation while preserving spatial information across heterogeneous medical imaging datasets. Optional MRI preprocessing includes N4 bias-field correction \cite{tustison2010n4}, and optional 3D volume rendering supports visual inspection of reconstructed structures. Automated test verifies image loading, mask import and resampling, source-grid volumetry, synchronization editing, and export. The software is designed exclusively for research annotation workflows and does not provide clinical decision support.

\section{Research impact statement}
Interactive Medical-SAM2 GUI has been used to create and revise source-aligned 3D labels in medical imaging research, including longitudinal analyses requiring assessment of segmentation continuity and volume changes across examinations. Combining its multi-label masks, object-wise masks, and volume reports supports dataset curation, inter-reader quality assurance, model training and validation, and longitudinal tumor-burden analysis. The local-first workflow allows protected medical images to remain within institutional computing environments, preserving image geometry and object identifiers. This is particularly relevant for cohort-scale annotation studies in which repeated file handling, inconsistent output geometry, and transfer of clinical images to external services may limit reproducibility and practical deployment.

This software workflow and core functionality were presented at the 20th Korean Brain Tumor Society Winter Meeting \cite{hong2026kbts}. A subsequent preliminary crossover study to compare the performance of this tool with manual segmentation using 3D Slicer for three meningioma MRI cases by four experienced neurosurgeons was presented at the 44th Annual Spring Meeting of the Korean Neurosurgical Society \cite{hong2026kns}. The semi-automated workflow showed comparable mean overlap accuracy (DSC, 0.923 vs. 0.917), lower mean surface distance (ASSD, 0.414 vs. 0.469 mm), reduced mean interaction time (-10.1\%), and lower inter-observer variability. As the differences in mean performance metrics were not statistically significant with only a few cases and readers included in the study, these findings should be interpreted as preliminary rather than as definitive clinical validation.

\section{AI usage disclosure}
OpenAI Codex (GPT-5) and GitHub Copilot were used for software implementation, debugging, and automated test generation. All AI-generated suggestions were reviewed, revised as needed, and validated by the authors through code review, automated testing, and manual GUI evaluation. Generative AI was not used to generate or analyze study data or scientific results. The authors retain responsibility for all scientific, architectural, and interface-design decisions.

\section{Conflict of interest}
The authors declare no competing interests.

\section*{Acknowledgements}
This development was supported by the National Research Foundation of Korea (NRF) grant (No.RS-2025-00517614). We thank the developers of Napari \cite{napari2025}, SimpleITK \cite{lowekamp2013simpleitk}, pydicom \cite{mason2024pydicom}, NiBabel \cite{brett2025nibabel}, PyVista \cite{sullivan2019pyvista}, SAM \cite{kirillov2023sam}, SAM2 \cite{ravi2025sam2}, and Medical-SAM2 \cite{zhu2024medical} for releasing open-source software and models.

\bibliographystyle{unsrt}
\bibliography{library}

\end{document}